\documentclass[letterpaper, 10 pt, conference]{ieeeconf}  % Comment this line out if you need a4paper

\IEEEoverridecommandlockouts                              % This command is only needed if 
\usepackage{graphicx} % for pdf, bitmapped graphics files
\usepackage{amsmath} % assumes amsmath package installed
\usepackage{amssymb}  % assumes amsmath package installed
\usepackage{booktabs}
\usepackage{arydshln}
\usepackage[compress]{cite}

\usepackage{comment}

\usepackage{caption}
\title{\LARGE \bf
\texttt{SARL:} Spatially-Aware Self-Supervised Representation Learning for Visuo-Tactile Perception
}

\author{Gurmeher Khurana*, Lan Wei* and Dandan Zhang
\thanks{*Equal Contribution.}
\thanks{Gurmeher Khurana, Lan Wei and Dandan Zhang are with the Department of Bioengineering, Imperial-X Initiative, Imperial College London, London, United Kingdom.  Corresponding: d.zhang17@imperial.ac.uk.}
}
\begin{document}

\maketitle
\thispagestyle{empty}
\pagestyle{empty}

%%%%%%%%%%%%%%%%%%%%%%%%%%%%%%%%%%%%%%%%%%%%%%%%%%%%%%%%%%%%%%%%%%%%%%%%%%%%%%%%
\begin{abstract}

% While vision provides global context, it often misses physical properties like texture and hardness; touch supplies these high-fidelity cues. 
% Fusing vision with touch mirrors human dexterity, motivating a learning approach that respects the spatial structure in multimodal data. This is especially true for hardware-fused visuo-tactile sensors, where visual and tactile data are intrinsically aligned at the pixel level, an ideal substrate for spatial objectives that is underserved by existing Self-Supervised Learning (SSL) frameworks, which are often misaligned with this need. 
Contact-rich robotic manipulation requires representations that encode local geometry. Vision provides global context but lacks direct measurements of properties such as texture and hardness, whereas touch supplies these cues. Modern visuo-tactile sensors capture both modalities in a single fused image, yielding intrinsically aligned inputs that are well suited to manipulation tasks requiring visual and tactile information.
Most self-supervised learning (SSL) frameworks, however, compress feature maps into a global vector, discarding spatial structure and misaligning with the needs of manipulation.
To address this, we propose \texttt{SARL}, a spatially-aware SSL framework that augments the Bootstrap Your Own Latent (BYOL) architecture with three map-level objectives, including Saliency Alignment (SAL), Patch-Prototype Distribution Alignment (PPDA), and Region Affinity Matching (RAM), to keep attentional focus, part composition, and geometric relations consistent across views. 
These losses act on intermediate feature maps, complementing the global objective.
SARL consistently outperforms nine SSL baselines across six downstream tasks with fused visual-tactile data.
On the geometry-sensitive edge-pose regression task, SARL achieves a Mean Absolute Error (MAE) of 0.3955, a 30\% relative improvement over the next-best SSL method (0.5682 MAE) and approaching the supervised upper bound. 
These findings indicate that, for fused visual–tactile data, the most effective signal is structured spatial equivariance, in which features vary predictably with object geometry, which enables more capable robotic perception.

\end{abstract}

%%%%%%%%%%%%%%%%%%%%%%%%%%%%%%%%%%%%%%%%%%%%%%%%%%%%%%%%%%%%%%%%%%%%%%%%%%%%%%%%
\section{Introduction}

Achieving human-level dexterity in robotic manipulation remains difficult, largely due to limited perceptual capabilities. Standard computer vision provides object recognition and scene context but lacks direct access to physical properties such as texture, hardness, and friction that are essential for stable interaction \cite{anytouch}. Humans overcome this limitation by integrating global visual cues with high-fidelity tactile feedback \cite{lin2023attention}, motivating tactile sensing as a critical complement in robotic systems.

To realise this complementary role of touch in practice, robotics has increasingly turned to vision-based tactile sensors (VBTSs), which treat touch as imaging: an internal camera observes a deformable, illuminated skin to recover contact geometry and related cues \cite{fan2025crystaltac}. As image-based signals, VBTS outputs integrate naturally with modern deep learning pipelines and have enabled strong tactile perception models \cite{fan2023tac,babadian2023fusion,gao2016deep}. This paradigm now supports a diverse family of tactile end-effectors for contact-rich manipulation \cite{zhang2024tacpalm}. While VBTSs provide rich local contact cues, they complement rather than replace global visual information, and robots often benefit from fusing both modalities \cite{}. However, using separate visual and tactile sensors introduces hardware redundancy and calibration challenges, motivating fused visuo–tactile sensing within a single device. We adopt the ViTacTip sensor \cite{fan2402vitactip}, whose transparent skin and biomimetic markers allow a single internal camera to capture the object's appearance and tactile deformation in a unified, intrinsically aligned image \cite{vitactip}. This fused pixel-space representation provides an ideal substrate for multimodal learning.
%fan2022graph, babadian2023fusion, 
\begin{figure}[!t]
\centering
\captionsetup{font=footnotesize,labelsep=period}
\includegraphics[width=0.8\columnwidth]{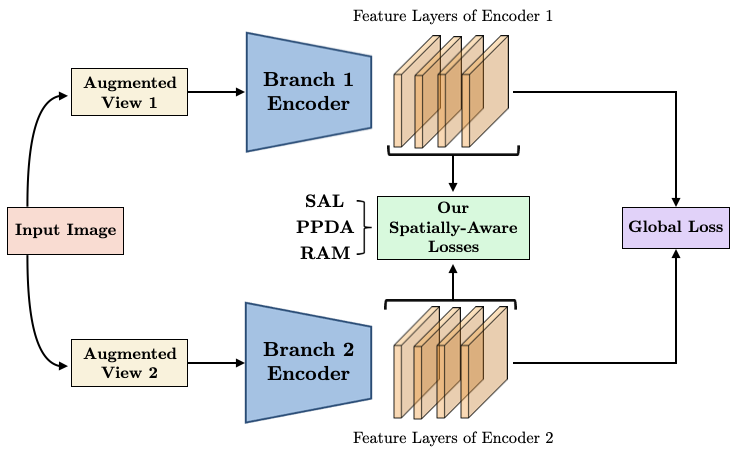}
\vspace{-0.15cm}
    \caption {Concept overview: \texttt{SARL} augments a Joint-Embedding Architecture with Spatially-Aware losses (SAL, PPDA and RAM) on intermediate feature layers, complementing a Global loss on final embeddings to preserve spatial and geometric cues typically removed by global pooling, yielding manipulation-ready representations.}
\label{fig:overview}
\vspace{-0.9cm}
\end{figure}

With fused visuo–tactile data available, a key challenge is learning representations that capture both modalities effectively. Self-supervised representation learning (SSL) \cite{ssl} has become central in robotics by enabling models to learn from interaction data without manual annotation, improving sample efficiency and generalisation \cite{zhang2024self}. However, most SSL approaches follow a joint-embedding architecture (JEA) \cite{jea}, as in SimCLR \cite{simclr} or Bootstrap Your Own Latent (BYOL) \cite{byol}, which use global pooling to enforce viewpoint and deformation invariance \cite{sslsurvey}. This produces globally invariant embeddings but discards spatial structure, limiting performance on geometry-sensitive tasks. Multimodal methods such as MViTac \cite{mvitactip} align modalities only at the global level, while spatially aware tactile methods like Sparsh \cite{sparsh} operate on a single modality. Thus, existing SSL methods cannot exploit the spatially aligned structure of fused visuo–tactile data.

To address this limitation, we propose \texttt{SARL} (Spatially-Aware self-supervised Representation Learning), a framework tailored to the spatial reasoning demands of contact-rich manipulation. \texttt{SARL} extends the non-contrastive BYOL architecture with three auxiliary losses applied to intermediate feature maps: Saliency Alignment (SAL), Patch-Prototype Distribution Alignment (PPDA), and Region Affinity Matching (RAM). These losses preserve attentional focus, semantic part structure, and geometric relationships across augmentations, enabling spatially structured multimodal representations. Our experiments show that this spatially aware learning objective yields substantial performance gains over existing SSL models.

The \textbf{main contributions} of this work are:
\begin{enumerate}
    \item We propose \texttt{SARL}, a spatially aware SSL framework for fused visuo–tactile data, which augments a non-contrastive baseline with three map-level losses (SAL, PPDA, RAM) to enforce consistency of attention, parts, and geometry.

    \item We benchmark \texttt{SARL} on six downstream tasks and show that it consistently outperforms nine state-of-the-art SSL methods, with particularly strong gains on geometry-sensitive tasks (e.g., a 30\% relative MAE reduction on edge-pose regression).

    \item We demonstrate robust transfer to four unseen tactile datasets and use ablations to confirm the complementary roles of the spatial losses and the advantage of fused multimodal representations over unimodal ones.
\end{enumerate}

\section{Related Work}

\subsection{Self-supervised Representation Learning Methods}

\begin{comment}
 The field of SSL has evolved from early heuristic pretext tasks, such as solving jigsaw puzzles \cite{noroozi2016unsupervised} or predicting image rotations \cite{imgrot}, to a more general principle of instance discrimination.    
\end{comment}

Modern SSL is dominated by several major paradigms. \emph{Contrastive learning methods}, such as SimCLR \cite{simclr} and MoCo \cite{moco}, learn by pulling representations of augmented `positive' views of an image together in an embedding space while pushing them apart from `negative' views of other images. These methods learn powerful discriminative features, but often require large batch sizes or memory banks. 
\emph{Distillation methods} like BYOL \cite{byol} and SimSiam \cite{simsiam} emerged, eliminating the need for negative pairs. They employ an asymmetric teacher-student architecture, where an `online' network is trained to predict the output of a `target' network whose weights are a slow-moving average of the online network's, preventing collapse through mechanisms like a momentum encoder or a stop-gradient operation. 

Other categories include \emph{information maximisation methods} (e.g., Barlow Twins \cite{barlow}, VICReg \cite{vicreg}), which prevent collapse by enforcing statistical properties like feature decorrelation, and \emph{clustering-based methods} (e.g., SwAV \cite{swav}), which use online clustering as a pretext task. Despite their different mechanisms, these dominant frameworks are designed to produce a single global feature vector, thus discarding the essential spatial information for dense prediction tasks.

\subsection{Representation Learning for Visuo-Tactile Perception}
The application of SSL to visuo-tactile perception is fundamentally constrained by the hardware configuration of the sensing system.     
The \emph{first category} of methods assumes physically independent sensors, such as an external camera and a fingertip-mounted tactile sensor. This separation necessitates a dual-encoder architecture where each modality is processed independently before fusion. For example, MViTac \cite{mvitactip} uses paired encoders with momentum updates and an InfoNCE objective, a contrastive loss, to align global latent spaces of vision and touch. Similarly, Touch-and-Go aligns \cite{touch} image and touch features via contrastive multiview coding, while UpViTaL \cite{upvital} relaxes the need for paired data by pre-training on unpaired datasets. 
Although consistent with the hardware design, these approaches perform global pooling per modality, which eliminates the spatial alignment necessary for contact localisation.

The \emph{second category} of methods focuses on unimodal representation learning from the output of a single VBTS. This setting naturally accommodates more spatially sensitive objectives, since the model operates directly on dense tactile images. The Sparsh framework \cite{sparsh} provides a benchmark of methods such as Masked Autoencoders (MAE) and DINO (self-distillation with no labels) on unimodal tactile data. Other approaches include UniT \cite{uniT}, which leverages a generative VQGAN backbone, and GS-DepthNet \cite{GelStereo}, targeting dense 3D contact geometry reconstruction from a stereo VBTS. While these methods can be spatially aware, their learning remains confined to a single modality.
A complementary line of work pursues sensor-agnostic generalisation. Frameworks such as T3 \cite{t3} and AnyTouch \cite{anytouch} aim to learn universal representations that transfer across diverse tactile sensors, but this emphasis on a “lowest common denominator’’ representation is distinct from our goal of fully exploiting the unique, high-fidelity information available in a single, hardware-fused visuo–tactile stream.

There is currently no self-supervised framework explicitly designed to learn from the dense, pixel-level visuo–tactile correspondences that arise when both modalities are intrinsically aligned in hardware, as in the ViTacTip sensor \cite{vitactip}. In particular, the field lacks an objective that simultaneously enforces cross-modal consistency and preserves the precise spatial structure required for dexterous, contact-rich manipulation. Addressing this gap is the focus of our proposed \texttt{SARL} framework.

\section{Methodology}
% This section details our proposed framework \texttt{SARL} to learn dense, structured representations from fused visuo-tactile data. We begin by formulating the problem, then present the \texttt{SARL} architecture, and detail its novel learning objectives.

\begin{figure*}[t!]
\centering
\captionsetup{font=footnotesize,labelsep=period}
\includegraphics[width=0.85\hsize]{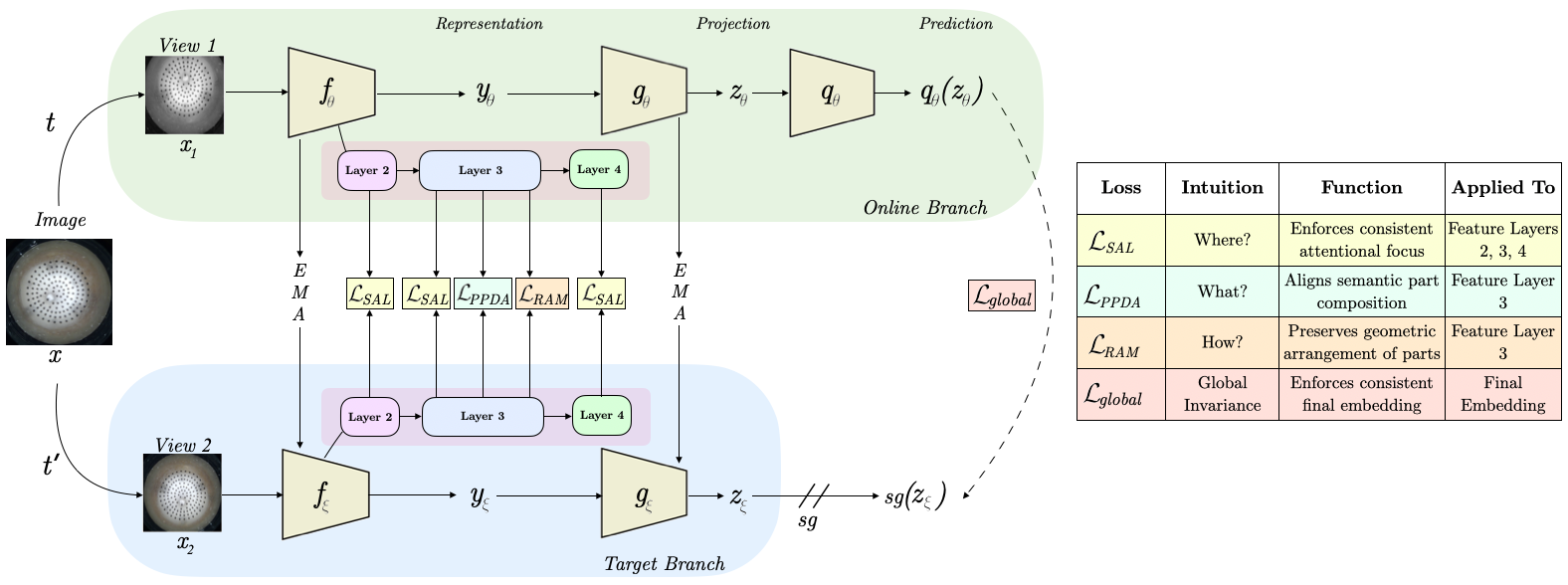}
\vspace{-0.15cm}
\caption[The \texttt{SARL} architecture.]
{\texttt{SARL} architecture. Input $x$ is augmented to $x_1,x_2$ and processed by parallel Online ($\theta$) and Target ($\xi$) branches. 
The global loss $\mathcal{L}_{\text{global}}$ trains $q_\theta$ to match $g_\xi$; Target weights are an EMA of Online and use stop-grad ($\mathrm{sg}$). 
From encoder features $f_\theta,f_\xi$, \texttt{SARL} adds spatial losses: $\mathcal{L}_{SAL}$ (Layers 2–4), $\mathcal{L}_{PPDA}$ (Layer 3 with a $7 \times 7$ grid and $K=32$ prototypes), and $\mathcal{L}_{RAM}$ (Layer 3 with a $6 \times 6$ grid).}
    % \caption [The \texttt{SARL} architecture.]
    % {The \texttt{SARL} architecture. An input image $x$ is transformed into two augmented views, $x_1$ and $x_2$. These views are processed by two parallel networks: the Online Branch (top, parametrised by $\theta$) and the Target Branch (bottom, parametrised by $\xi$). The global objective, $\mathcal{L}_{\text{global}}$, is computed by training the online network's predictor ($q_{\theta}$) to match the target network's projection ($g_{\xi}$). The target network's weights are a slow-moving average of the online network's (updated via EMA) and its gradients are stopped ($\mathrm{sg}$). \texttt{SARL} extends this by extracting intermediate feature maps from both encoders ($f_{\theta}$ and $f_{\xi}$) to compute three additional spatially-aware losses: Saliency Alignment ($\mathcal{L}_{SAL}$; Layers 2, 3, \& 4), Patch-Prototype Distribution Alignment ($\mathcal{L}_{PPDA}$; Layer 3 with a $7 \times 7$ grid and $K=32$ prototypes), and Region Affinity Matching ($\mathcal{L}_{RAM}$; Layer 3 with a $6 \times 6$ grid).}
\label{fig:sarl_architecture}
\vspace{-0.6cm}
\end{figure*}

\subsection{Problem Formulation \& Hypothesis}

The objective is to learn an encoder $f: \mathcal{X} \to \mathbb{R}^d$ that maps a high-dimensional visuo-tactile image $x \in \mathcal{X}$ to a low-dimensional, semantically meaningful and spatially-aware feature representation, without access to human-provided labels. Following the standard SSL paradigm, we assume a distribution of stochastic image augmentations $\mathcal{T}$. For any given image $x$, two distinct augmented views, $x_1 = t(x)$ and $x_2 = t'(x)$ with $t, t' \sim \mathcal{T}$, are generated.

%Conventional SSL encourages the encoder $f$ to learn augmentation-invariant global embeddings. While suitable for category recognition, this design compresses fine-grained spatial cues into a single vector and weakens dense prediction. Manipulation tasks, including pose estimation and texture identification, instead require detailed representations of object geometry and part-level semantics.

Conventional SSL encourages the encoder $f$ to learn augmentation-invariant global embeddings. Although this is effective for high-level classification tasks, it creates a bottleneck that compresses spatial and structural details into a single vector. 
For robotic manipulation, tasks such as pose estimation and texture identification demand representations that retain local geometry and part-level semantics.
This motivates our central hypothesis: an SSL framework that explicitly enforces spatial and semantic consistency at the feature level will produce representations that yield superior performance on fine-grained manipulation tasks compared to methods that only enforce global feature invariance.

This hypothesis is based on the premise that richer representations can be learned by integrating three key inductive biases directly into the training objective. We propose that a model's performance will improve if it is trained to:
% \vspace{-0.2cm}
\begin{itemize}
    \item Maintain a consistent focus on salient object regions across views, learning what parts of an object are important.
    \item Ensure that local features corresponding to the same object part are semantically consistent, promoting a detailed, part-level understanding.
    \item Preserve the geometric relationships among different regions of an object, encouraging the model to learn the object's structure.
\end{itemize}

We instantiate this hypothesis with three spatial objectives that act on intermediate feature maps: \textbf{Saliency Alignment (SAL)} for attentional alignment, \textbf{Patch-Prototype Distribution Alignment (PPDA)} for semantic part agreement, and \textbf{Region Affinity Matching (RAM)} for pairwise geometric consistency. These constraints complement the global objective and yield spatially aware, semantically rich representations for downstream manipulation tasks.

\subsection{\texttt{SARL} Architectural Design}

\texttt{SARL} builds upon the architectural principles of BYOL\cite{byol}. We chose BYOL as our foundation because its joint-embedding predictive architecture, a non-contrastive approach, provides a stable and effective mechanism for preventing representational collapse without requiring negative sampling.

The overall \texttt{SARL} framework employs two neural networks: an online network with parameters $\theta$ and a target network with parameters $\xi$. Each network comprises an encoder $f$ that produces a representation and a projector $g$ that maps the representation into a latent embedding space. Additionally, the online network includes a predictor $q$ that attempts to predict the target embedding. To prevent collapse, the learning dynamic is stabilised by two key mechanisms. First, the target network's weights are not updated via backpropagation but are instead an exponential moving average (EMA) of the online network's weights, providing a stable, slowly-evolving learning target, updated after each training step $t$ with a decay rate $\mu$:
\vspace{-0.15cm}
\begin{equation}
    \xi_{t} \leftarrow \mu \xi_{t-1} + (1-\mu) \theta_{t}
    \vspace{-0.15cm}
\end{equation}
Second, and crucially, all outputs from the target network are treated as fixed learning targets. A stop-gradient (sg) is applied to them, both the final projections for the global loss and the intermediate feature maps for the spatial losses, to ensure no gradients propagate back through the target network.

The \texttt{SARL} objective begins with a global invariance loss, $\mathcal{L}_{\text{global}}$, inherited from BYOL \cite{byol}. This loss minimises the mean squared error between the L2-normalised prediction from the online network and the projection from the target network. For two augmented views $x_1$ and $x_2$, the loss is symmetrised:
% This momentum update yields a stable target network whose outputs serve as regression targets for the online network, mitigating collapse by slowly bootstrapping from previous outputs \cite{byol}.
% The training objective is composed of a global invariance loss, inherited from BYOL \cite{byol}, and three novel spatially-aware losses. 
% The global loss, $\mathcal{L}_{\text{global}}$  minimizes the mean squared error between the L2-normalized prediction from the online network and the projection from the target network. A stop-gradient (sg) is applied to the target projection to prevent collapse. For two augmented views $x_1$ and $x_2$, the loss is symmetrized:
% \vspace{-0.15cm}
\begin{equation}
\begin{split}
    \mathcal{L}_{\text{global}} \;=\; & \left\|\, \bar{q}_{\theta}(z_{\theta}(x_1)) - \mathrm{sg}[\bar{z}_{\xi}(x_2)] \right\|_2^2 \\
    & + \left\|\, \bar{q}_{\theta}(z_{\theta}(x_2)) - \mathrm{sg}[\bar{z}_{\xi}(x_1)] \right\|_2^2
\end{split}
\vspace{-0.15cm}
\end{equation}
Here, $\bar{q}_{\theta}(z_{\theta})$ and $\bar{z}_{\xi}$ represent the L2-normalised predictor and target projector outputs, respectively.

Although $\mathcal{L}_{\text{global}}$ learns powerful invariant features, it discards spatial information by design. The core innovation of \texttt{SARL} is to augment this objective with three auxiliary losses that operate directly on the intermediate feature maps from the online and target encoders, $F_{\theta}(x) \in \mathbb{R}^{H\times W \times C}$ and $F_{\xi}(x) \in \mathbb{R}^{H\times W \times C}$, respectively. This architecture, depicted in Fig.~\ref{fig:sarl_architecture}, allows the model to learn from both global and local consistency signals simultaneously.

\subsection{Spatially-Aware Learning Objectives}

The three auxiliary losses form a synergistic system that addresses different aspects of spatial awareness. SAL answers `where to look', PPDA answers `what parts are there', and RAM answers `how are the parts arranged'.

\subsubsection{SAL [``where'']}

The SAL loss is designed to ensure the model consistently focuses on the same salient object regions across different views, promoting a stable attentional mechanism. For a given feature map $F(x) \in \mathbb{R}^{H \times W \times C}$, we first compute its saliency map $S(x) \in \mathbb{R}^{H \times W}$ as the channel-wise L1 norm (sum of absolute values) of the feature activations at each spatial location $(i,j)$.

The geometric augmentation that transforms an image into two views, $x_1$ and $x_2$, defines an affine warp that maps coordinates from view 1 to view 2 via a function $\pi(i,j)$. Because random cropping can cause some pixels in one view to have no corresponding location in the other, we define a valid overlap mask, $M$, which contains only the coordinates $(i,j)$ in $x_1$ for which $\pi(i,j)$ is within the spatial bounds of $x_2$. Before comparison, both saliency maps are L2-normalised. The loss then minimises the mean squared error (MSE) exclusively over the valid, overlapping regions:
\vspace{-0.15cm}
\begin{equation}
\mathcal{L}_{SAL} = \frac{1}{|M|} \sum_{(i,j) \in M} \big( S_{\theta}(x_1)[i,j] - S_{\xi}(x_2)[\pi(i,j)] \big)^2
% \vspace{-0.15cm}
\end{equation}
In essence, this loss encourages the model to focus on the same physical locations in both augmented views, forcing attention to remain consistent under geometric transformations.
This loss is applied across layers 2, 3, and 4 of the encoder to enforce hierarchical attentional consistency. For each layer, the coordinate mapping $\pi$ is scaled proportionally to the spatial resolution of that layer’s feature map, ensuring alignment between saliency maps of different sizes.
If the feature maps for $x_1$ and $x_2$ have mismatched spatial dimensions due to data augmentations,
the target saliency map $S_{\xi}(x_2)$ is resized using bilinear interpolation to match the online map, ensuring that $\pi(i,j)$ indexes a valid location and enabling direct pixelwise comparison.

\subsubsection{PPDA [``what'']}

The PPDA loss enforces local semantic consistency by ensuring that two views of an object are composed of the same distribution of semantic “parts’’ or “textures’’. We introduce a set of $K$ learnable prototype vectors $\{p_k\}_{k=1}^K \subset \mathbb{R}^C$, where the dimension $C$ matches the channel count of the feature map (256 for Layer 3). Intuitively, each prototype $p_k$ represents a latent “part type’’ (for example, a particular edge, corner, or texture pattern) that can appear at different spatial locations. Both the prototypes and the local feature vectors (patches), $v_i$, extracted from the feature map are L2-normalised so that their inner product corresponds to cosine similarity. Each patch is then softly assigned to the prototypes using a softmax function, where
$q_i(k) \propto \exp(\langle v_i, p_k \rangle / \tau)$ and the values are normalised over $k$ to form a probability distribution. Here, $\langle v_i, p_k \rangle$ measures the match between patch $i$ and prototype $k$, and the temperature $\tau$ controls how sharp the assignments are.

%which converts the similarity scores between the patch feature $v_i$ and each prototype $p_k$ into a probability distribution over prototypes. Here, $\langle v_i, p_k \rangle$ measures how well patch $i$ matches prototype $k$, and the temperature $\tau$ controls the sharpness of this distribution: a lower $\tau$ yields more peaked (harder) assignments, whereas a higher $\tau$ produces softer, more uniform assignments. After normalisation, $q_i(k)$ can be interpreted as the probability that patch $i$ belongs to prototype $k$.

The aggregated distribution for the entire image, $Q(x) \in \mathbb{R}^K$, is computed as the spatial average of all the patch-level soft assignments,
$Q(x)(k) = \frac{1}{HW}\sum_i q_i(k)$, where $H$ and $W$ are the height and width of the feature map and $HW$ is the total number of patches. Thus, $Q(x)$ describes how frequently each prototype (i.e., part type) appears across the whole image, forming a probability distribution that sums to one. The loss then minimises the symmetric Kullback–Leibler (KL) divergence between the prototype distributions of the online and target views:
\vspace{-0.15cm}
\begin{equation}
\begin{aligned}
\mathcal{L}_{PPDA} &= D_{KL}\!\big(Q_{\theta}(x_1)\,\|\,Q_{\xi}(x_2)\big) \\
&\quad + D_{KL}\!\big(Q_{\xi}(x_2)\,\|\,Q_{\theta}(x_1)\big)
\end{aligned}
\vspace{-0.15cm}
\end{equation}
so that both views are encouraged to exhibit the same global mixture of parts. This objective is applied at Layer 3 of the encoder, which provides an effective trade-off between semantic richness (prototypes capture meaningful parts) and spatial resolution (patches still retain local detail), making it well suited for representing object parts relevant to manipulation.

\subsubsection{RAM [``how'']}

The RAM loss preserves the geometric structure of an object by ensuring the pairwise relationships between different feature regions remain constant across views. We first partition the feature map into a grid of regions (e.g., $6 \times 6 = 36$ regions). For each region, we extract its feature vector and apply L2-normalisation. The affinity between any two region vectors, $v_i$ and $v_j$, is then explicitly defined as their cosine distance: $a_{ij} = 1 - \frac{v_i \cdot v_j}{\|v_i\| \|v_j\|}$.

This process is used to compute a complete region affinity matrix, $A(x)$, for each view, which captures the pairwise affinities between all regions in the grid. The loss then minimises the mean squared error between the affinity matrix from the online network, $A_{\theta}(x_1)$, and the corresponding matrix from the target network, $A_{\xi}(x_2)$.
\vspace{-0.15cm}
\begin{equation}
\mathcal{L}_{RAM} = \frac{1}{P^2}\sum_{i=1}^{P}\sum_{j=1}^{P} \big( A_{\theta}(x_1)_{ij} - A_{\xi}(x_2)_{\pi(i)\pi(j)} \big)^2
\vspace{-0.15cm}
\end{equation}
Here, $P$ is the total number of regions in the grid (e.g., 36), and $\pi$ maps the region indices from one view to the other. Like PPDA, this loss is applied at Layer 3, where the features are sufficiently abstract to capture the object's internal geometry while retaining high spatial resolution.

% The Region Affinity Matching (RAM) loss preserves the geometric structure of an object by ensuring that the pairwise relationships (affinities) between different feature regions remain constant across views. 
% For a sampled pair of feature vectors $(v_i, v_j)$ from the online feature map $F_{\theta}(x_1)$, we compute their cosine similarity $a_{ij}^{(1)}$. We then compute the affinity $a_{\pi(i)\pi(j)}^{(2)}$ for the corresponding pair in the target feature map $F_{\xi}(x_2)$. The loss minimizes the squared difference between these affinities, averaged over a large number of randomly sampled pairs:
% \begin{equation}
% \mathcal{L}_{RAM} = \mathbb{E}_{(i,j)}[(a_{ij}^{(1)} - a_{\pi(i)\pi(j)}^{(2)})^2]
% \end{equation}
% Like PPDA, this loss is applied at Layer 3, where the features are sufficiently abstract to capture the object's internal geometry while retaining high spatial resolution.
\subsubsection{Combined Objective Function}

The full \texttt{SARL} training objective is a weighted sum of the global loss and the three spatially-aware auxiliary losses:
\vspace{-0.15cm}
\begin{equation}
\begin{aligned}
\mathcal{L}_{\texttt{SARL}} &= \mathcal{L}_{global} + \lambda_{SAL}\mathcal{L}_{SAL} \\
&\quad + \lambda_{PPDA}\mathcal{L}_{PPDA} + \lambda_{RAM}\mathcal{L}_{RAM}
\end{aligned}
\vspace{-0.15cm}
\end{equation}

\begin{figure*}[t!]
\centering
\captionsetup{font=footnotesize,labelsep=period}
\includegraphics[width=0.9\textwidth]{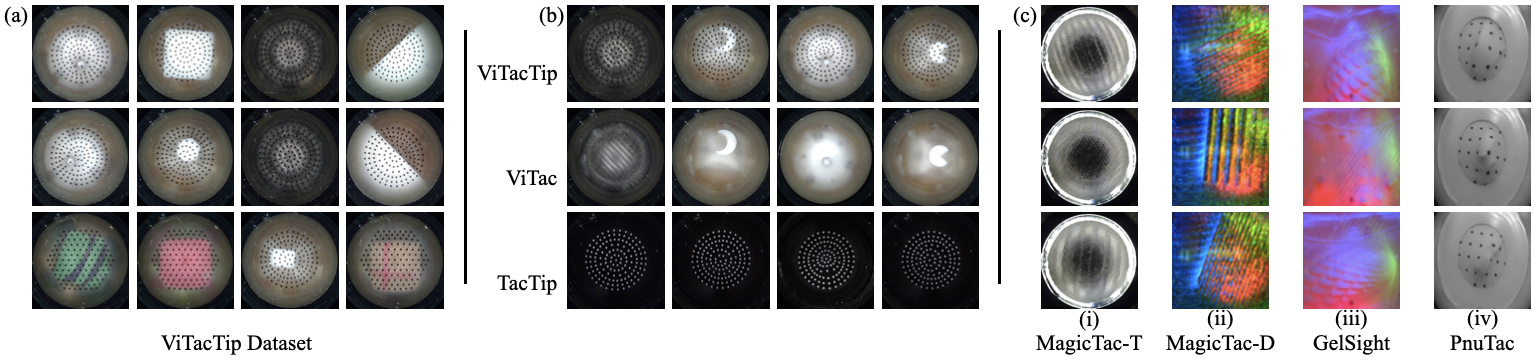}
\vspace{-0.2cm}
\caption{Overview of the datasets used for pre-training, evaluation, and transfer learning. 
\textbf{(a)} Sample images from the primary ViTacTip dataset, showcasing the fused visuo-tactile data across several downstream tasks. 
\textbf{(b)} The three data modalities used in our ablation studies: the fused multimodal ViTacTip data (top row), its visual-only ViTac counterpart (middle row), and its marker-only TacTip counterpart (bottom row). 
\textbf{(c)} Sample images from the four unseen datasets used for the generalizability evaluation: (i) MagicTac-T, (ii) MagicTac-D, (iii) GelSight, and (iv) PnuTac.}
\label{fig:dataset_overview}
\vspace{-0.5cm}
\end{figure*}

These auxiliary losses are complementary: SAL enforces consistent attentional focus (\textit{where}), PPDA aligns the composition of semantic parts (\textit{what}), and RAM preserves the geometric relationships between them. We use balancing coefficients $\lambda_{SAL}$, $\lambda_{PPDA}$, and $\lambda_{RAM}$. Their values and the tuning process are detailed in Section~\ref{sss:implementation}. 
This combined objective trains the encoder to produce representations that are simultaneously globally invariant to nuisance factors and locally structured to preserve task-relevant geometric and semantic information.

\section{Experiments}

\subsection{Dataset}

% \begin{figure*}[t!]
% \centering
% \captionsetup{font=footnotesize,labelsep=period}
% \includegraphics[width=\textwidth]{dataset_overview.png}
% \vspace{-0.5cm}
% \caption{Overview of the datasets used for pre-training, evaluation, and transfer learning. 
% \textbf{(a)} Sample images from the primary ViTacTip dataset, showcasing the fused visuo-tactile data across several downstream tasks. 
% \textbf{(b)} The three data modalities used in our ablation studies: (i) the fused multimodal ViTacTip data, (ii) its visual-only ViTac counterpart, and (iii) its tactile-only TacTip counterpart. 
% \textbf{(c)} Sample images from the four unseen datasets used for the generalizability evaluation: (i) MagicTac-T, (ii) MagicTac-D, (iii) GelSight, and (iv) PnuTac.}
% \label{fig:dataset_overview}
% \vspace{-0.1cm}
% \end{figure*}

We use the ViTacTip dataset for pre-training and evaluation (Fig.~\ref{fig:dataset_overview} (a)). 
To support reproducibility, we will publicly release the full dataset.
% We use the ViTacTip dataset \cite{vitactip} for pre-training and evaluation (Fig.~\ref{fig:dataset_overview} (a)). 
% It is collected using a vision-based tactile sensor that captures a single, fused image containing both external visual appearance and internal tactile information from marker displacement. 
The dataset spans six robotic perception tasks: shape classification, grating identification, multitask classification (material, hardness, texture), force regression, edge pose regression, and contact point detection. For SSL pre-training, we create a large, aggregated dataset by combining only the training and validation images from all tasks while holding out the corresponding test sets. 

For downstream evaluation, we use the original task-specific datasets with their labels, partitioned into disjoint training (70\%), validation (15\%), and test (15\%) splits, with final linear probe performance reported on the held-out test split. To assess generalizability, four additional visuo-tactile datasets were used to evaluate transfer learning. (Fig.~\ref{fig:dataset_overview} (c)). These include two in-house datasets from the MagicTac sensor family which is produced using the multimaterial additive manufacturing technique \cite{magictac}. The are MagicTac-T, which uses the same base as TacTip, and MagicTac-D, which is smaller and uses the same base as DIGIT. We also benchmarked two open source datasets from the University of Cardiff: the GelSight dataset, containing 2000 images for each of 9 different tool textures, and the PnuTac dataset, for which we used the subset of 11 hand tool classes (250 images each) \cite{pnutac}. Each of these transfer learning datasets was partitioned into training (60\%), validation (20\%), and test (20\%) splits. To quantify the benefit of multimodal visuo-tactile data, our ablation studies also compare performance on the fused ViTacTip data against its unimodal counterparts: TacTip, which only employs internal pins/markers and ViTac, which removes internal markers and leverages a transparent, see-through skin to prioritize direct visual observation of the contact patch (Fig.~\ref{fig:dataset_overview} (b)).

\subsection{Evaluation Metrics}
To rigorously assess the quality of learned representations, a multifaceted evaluation protocol was employed. The \emph{primary method} is linear probing, where the pre-trained encoder is frozen, and a single linear layer is trained on top of its features for each downstream task. For classification tasks, we report Top-1 (the highest probability prediction is correct) and Top-5 accuracy (the correct class is within the five highest probability predictions). For regression tasks, we report mean absolute error (MAE) with units defined by the task's coordinate frame. For contact point estimation and force regression, MAE is measured in millimetres (mm) for the $P_x, P_y, P_z$ and $F_x, F_y, F_z$ coordinates on the sensor's surface. For edge pose regression, MAE is measured in mm for the horizontal distance ($X$) and press depth ($Z$), and in degrees for the angle of rotation ($\theta$).

The \emph{second protocol} is transfer learning, which measures the generalizability of the representations. Here, the pre-trained encoder is fine-tuned on four unseen visuo-tactile datasets, testing its ability to adapt to new domains, and we report performance using Top-1 and Top-5 accuracy. 
% Finally, a series of ablation studies were conducted to validate the framework's design. These include component ablations, which systematically remove individual spatial loss functions to quantify their specific contribution, and modality ablations, which compare models trained on fused multimodal data against their unimodal (vision-only or tactile-only) counterparts to demonstrate the benefit of data fusion.

\begin{table*}[t!]
\centering
\captionsetup{font=footnotesize,labelsep=period}
\caption{Linear probe performance across all downstream tasks. For this protocol, the pre-trained encoder is frozen. The SSL pre-training pool consists of all unlabeled images from the train and validation splits. All inputs are 224$\times$224. Metrics are Top-1/Top-5 accuracy (\%) for classification and average MAE for regression (Force/Contact Point in mm; Edge Pose in mm and degrees).}
\label{tab:combined}
\footnotesize
\resizebox{\linewidth}{!}{
\begin{tabular}{lccccccccc}
\hline
& \multicolumn{6}{c}{\textbf{Classification Tasks}} & \multicolumn{3}{c}{\textbf{Regression Tasks}} \\
\cmidrule(lr){2-7}\cmidrule(lr){8-10}
& \multicolumn{2}{c}{\textbf{Shape}} & \multicolumn{2}{c}{\textbf{Grating}} & \multicolumn{2}{c}{\textbf{Material}} & \textbf{Force} & \textbf{Contact Point} & \textbf{Edge Pose} \\
\cmidrule(lr){2-3}\cmidrule(lr){4-5}\cmidrule(lr){6-7}\cmidrule(lr){8-8}\cmidrule(lr){9-9}\cmidrule(lr){10-10}
\textbf{Method} & \textbf{Top-1 (}\(\uparrow\)\textbf{)} & \textbf{Top-5 (}\(\uparrow\)\textbf{)} & \textbf{Top-1 (}\(\uparrow\)\textbf{)} & \textbf{Top-5 (}\(\uparrow\)\textbf{)} & \textbf{Top-1 (}\(\uparrow\)\textbf{)} & \textbf{Top-5 (}\(\uparrow\)\textbf{)} & \textbf{Avg.\ MAE (}\(\downarrow\)\textbf{)} & \textbf{Avg.\ MAE (}\(\downarrow\)\textbf{)} & \textbf{Avg.\ MAE (}\(\downarrow\)\textbf{)} \\
\hline
Supervised           & 96.75 & 100.00 & 100.00 & 100.00 & 99.77 & 100.00 & 0.0300 & 0.0828 & 0.3547 \\
\hdashline
SimCLR               & \underline{88.25} & \textbf{100.00} & \underline{93.52} & \textbf{100.00} & \underline{99.03} & 99.74 & \textbf{0.0529} & \underline{0.1977} & 0.6379 \\
BYOL                 & 82.67 & \underline{99.94} & 83.80 & \textbf{100.00} & 98.36 & 99.66 & 0.1041 & 0.3292 & 0.9529 \\
SimSiam              & 81.59 & \underline{99.94} & 85.33 & \textbf{100.00} & 97.50 & 99.68 & 0.1016 & 0.3334 & \underline{0.5682} \\
Barlow Twins         & \underline{86.22} & \textbf{100.00} & 89.90 & \textbf{100.00} & 98.41 & \underline{99.76} & \underline{0.0746} & \underline{0.2115} & \underline{0.5836} \\
VICReg               & 85.14 & \textbf{100.00} & \underline{90.86} & \textbf{100.00} & 98.28 & \underline{99.76} & 0.1736 & 1.3293 & 0.9372 \\
SwaV                 & 80.38 & 98.92 & 82.52 & \textbf{100.00} & \underline{98.70} & 99.68 & 0.0779 & 0.4042 & 0.8167 \\
MINC                 & 84.19 & \textbf{100.00} & 89.71 & \textbf{100.00} & 98.54 & 99.74 & 0.1089 & 0.3164 & 0.6072 \\
CPLearn              & 74.29 & 94.98 & 80.19 & \underline{99.80} & 96.05 & 99.55 & 0.1093 & 0.4202 & 1.0722 \\
DenseCL              & 85.87 & \underline{99.25} & 89.71 & \textbf{100.00} & 98.14 & \underline{99.76} & 0.0900 & 0.3221 & 0.6200 \\
\texttt{SARL} (ours) & \textbf{92.51} & \textbf{100.00} & \textbf{99.61} & \textbf{100.00} & \textbf{99.27} & \textbf{99.77} & \underline{0.0617} & \textbf{0.1804} & \textbf{0.3955} \\
\hline
\end{tabular}}
\vspace{2pt}
{\footnotesize Best result in each column is \textbf{bold}; 2nd and 3rd are \underline{underlined}. The Supervised baseline is a ResNet-18 and is included as a reference only (not ranked).}
\vspace{-0.5cm}
\end{table*}

\subsection{Experimental Settings}
\subsubsection{Implementation Details}\label{sss:implementation}

The encoder for both the online and target branches is a ResNet-18 \cite{resnet}, outputting a 512-dimensional representation from its final average pooling layer. This is mapped to a 256-dim latent space via a 2-layer MLP projection head. The online branch includes an additional 2-layer MLP prediction head, which also outputs a 256-dim vector. Our auxiliary losses are applied to intermediate feature maps: SAL is computed on layers 2, 3, and 4, while PPDA and RAM are computed on layer 3. The balancing coefficients for these losses were determined via grid search on a shape classification proxy task and set to $\lambda_{SAL}=0.10$, $\lambda_{PPDA}=0.05$, and $\lambda_{RAM}=0.02$. The PPDA module uses a 7$\times$7 grid, 32 prototypes, and a temperature of 0.1, while the RAM module uses a 6$\times$6 grid. The target network's parameters are updated via an EMA with a constant momentum of 0.996. We use AdamW, with a base learning rate of $1\times10^{-3}$, weight decay of $1\times10^{-4}$, and default values for moment estimates ($\beta_1 = 0.9$, $\beta_2 = 0.999$) for $100$ epochs using a batch size of $256$ on a single NVIDIA A100-SXM4-80GB GPU.

For downstream tasks, the pre-trained encoder was frozen, and a single linear layer was trained on top of the 512-dimensional backbone representation. For regression tasks, this head was trained for 200 epochs using AdamW with a learning rate of 0.01. For classification tasks, it was trained for 100 epochs using SGD with a learning rate of 0.02 and momentum of 0.9.

\subsubsection{Image Augmentations}

Our augmentation pipeline consists of several sequential operations to generate two correlated views for SSL pre-training. First, a random region is cropped, with its area uniformly sampled between 20\% and 100\% of the original image, and resized to 224$\times$224 pixels. This is followed by a horizontal flip with a 50\% probability. A colour jittering transformation is then applied with 80\% probability, randomly altering brightness, contrast, saturation, and hue. Subsequently, the image is converted to grayscale with 20\% probability. Finally, a Gaussian blur with a 23$\times$23 kernel is applied. All images are then normalised.

\subsubsection{Baseline Methods Implementation}

We benchmark \texttt{SARL} against nine state-of-the-art SSL methods: SimCLR \cite{simclr}, BYOL \cite{byol}, SimSiam \cite{simsiam}, Barlow Twins \cite{barlow}, VICReg \cite{vicreg}, SWAV \cite{swav}, MINC \cite{minc}, CPLearn \cite{cplearn}, and DenseCL \cite{densecl}. For fair comparison, all baselines use a ResNet-18 \cite{resnet} backbone and are trained with a batch size of 256 for 100 epochs on the unlabeled aggregated ViTacTip dataset, following the exact implementation details from their original papers. We also include a fully supervised ResNet-18 \cite{resnet} trained from scratch on the labelled downstream tasks, serving as an upper-bound reference.

% \begin{table*}[!t]
% \centering
% \captionsetup{font=footnotesize,labelsep=period}
% \caption{Transfer learning from ViTacTip on classification tasks across different datasets}
% \label{tab:transfer_linearprobe}
% \small
% \renewcommand{\arraystretch}{1.25}
% \begin{tabular}{l*{8}{c}}
% \hline
% & \multicolumn{2}{c}{\textbf{MagicTac-T}} &
%   \multicolumn{2}{c}{\textbf{MagicTac-D}} &
%   \multicolumn{2}{c}{\textbf{GelSight}} &
%   \multicolumn{2}{c}{\textbf{PnuTac}} \\
% \cmidrule(lr){2-3}\cmidrule(lr){4-5}\cmidrule(lr){6-7}\cmidrule(lr){8-9}
% \textbf{Method} &
% \textbf{Top-1} & \textbf{Top-5} &
% \textbf{Top-1} & \textbf{Top-5} &
% \textbf{Top-1} & \textbf{Top-5} &
% \textbf{Top-1} & \textbf{Top-5} \\
% \hline
% SimCLR         & 80.19 & 95.38 & 77.89 & \underline{95.24} & 85.89 & 94.54 & 79.65 & 93.78 \\
% BYOL           & 87.04 & 98.61 & 69.58 & 91.05 & 88.46 & 96.78 & 88.08 & \textbf{97.57} \\
% Barlow Twins   & \underline{90.85} & \underline{98.86} & \underline{78.38} & 94.24 & \textbf{92.30} & \textbf{99.12} & \underline{88.98} & 97.23 \\
% \texttt{SARL} (ours)  & \textbf{91.61} & \textbf{98.88} & \textbf{80.48} & \textbf{96.40} & \underline{90.24} & \underline{98.54} & \textbf{91.00} & \underline{97.32} \\
% \hline
% \end{tabular}

% \vspace{2pt}
% {\footnotesize Best result in each column is \textbf{bold}; second-best is \underline{underlined}.}
% \end{table*}

\section{Result \& Analysis}

\subsection{Linear Probing}

Across the six downstream tasks, \texttt{SARL} consistently outperformed all nine SSL baselines, validating the effectiveness of its spatially-aware objectives. Table~\ref{tab:combined} summarises the linear probing performance, where \texttt{SARL} achieves the highest Top-1 accuracy in all three classification tasks and the lowest average MAE in two of the three regression tasks. The largest gains appear on tasks that require fine-grained geometric reasoning, which provides strong evidence for our central hypothesis. 

\subsubsection{Regression Tasks}
\begin{figure}[t!]
\centering
\captionsetup{font=footnotesize,labelsep=period}
\includegraphics[width=0.9\columnwidth]{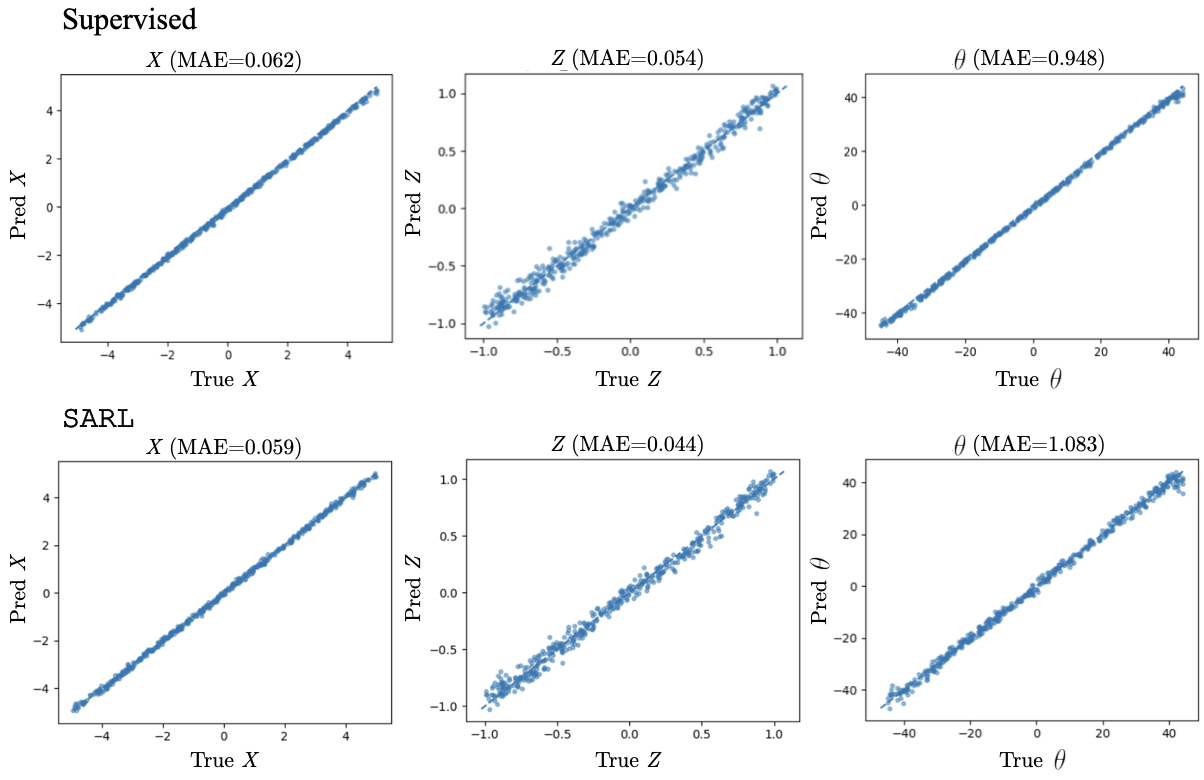}
\vspace{-0.4cm}
    \caption {Linear probe performance on the Edge Pose Regression task, comparing \texttt{SARL} (bottom row) against the fully supervised baseline (top row). Each plot shows the predicted vs. true values for the three pose components: horizontal distance ($X$ in mm), press depth ($Z$ in mm), and angle of rotation ($\theta$ in degrees).}
\label{fig:edge_peraxis}
\vspace{-0.6cm}
\end{figure}

For the \emph{edge pose regression task}, \texttt{SARL} achieved an average MAE of 0.3955, a 30.4\% relative improvement over the next-best SSL method, SimSiam (0.5682 MAE). This result underscores the benefit of \texttt{SARL}'s auxiliary losses (SAL, PPDA, and RAM), which explicitly preserve the local structural information that global SSL methods are designed to discard. A particularly noteworthy finding is observed in the per-axis results for this task, illustrated in Fig.~\ref{fig:edge_peraxis}. While the supervised baseline is superior on average due to the rotational component ($\theta$), \texttt{SARL}'s learned features are more effective for predicting the translational components, achieving a lower MAE on both the X-axis (0.0596 vs. 0.0621) and Z-axis (0.0444 vs. 0.0539). This suggests that self-supervision can act as a powerful regularizer, forcing the model to learn more general and robust geometric features than a supervised approach that may overfit. A clear performance trade-off emerges when comparing this to the \emph{force regression task}. Here, the global method SimCLR performs best, likely because predicting net force relies on a low-frequency, holistic view of the sensor's deformation, for which globally-invariant features are well-suited. \texttt{SARL}'s strong second-place performance indicates its global features remain robust, but the contrast between these two tasks highlights the importance of matching the inductive bias of the learning algorithm to the demands of the specific manipulation task.

\subsubsection{Classification Tasks}
In classification, \texttt{SARL}'s spatially-aware features also proved superior. For the \emph{shape classification task}, \texttt{SARL} (92.51\% Top-1) reduces the error rate by 36.3\% compared to the second-best SSL method. This advantage is even more pronounced on tasks requiring fine-grained texture discrimination. On the grating identification task, \texttt{SARL} achieves 99.61\% accuracy, nearly matching the supervised upper bound. Similarly, it secures top performance on material classification at 99.27\%. These results demonstrate a clear advantage for spatially-aware self-supervision on fused visuo-tactile data. By enforcing local consistency, \texttt{SARL} learns representations that are not only superior for geometrically-sensitive tasks but also highly competitive for global classification, validating our core hypothesis and highlighting the potential for SSL to even surpass supervised methods on specific, dense prediction sub-tasks.

\subsection{Transfer Learning}

To assess the robustness and generalisation of the learned representations, we fine-tuned the pre-trained encoders of SimCLR, BYOL, Barlow Twins, and \texttt{SARL} on four unseen visuo-tactile datasets: MagicTac-T, MagicTac-D, GelSight, and PnuTac. Given the small size of these datasets, containing approximately 2500-5500 images each, the fine-tuning was conducted for 60 epochs. The protocol involved unfreezing only the final residual block (layer 4) of the encoder, while the classifier head was trained from scratch. All models were trained with the AdamW optimizer using a learning rate of $1\times 10^{-2}$, which was decayed using a cosine annealing schedule.

For a consistent evaluation, all models used a multipool readout head concatenating features from frozen layer 3 and trainable layer 4. This protocol directly tests the utility of mid-level features learned during pre-training and is particularly synergistic with \texttt{SARL}, as two spatial losses are applied at layer 3. The results presented in Table~\ref{tab:transfer_linearprobe} confirmed \texttt{SARL}'s strong generalization; it achieved the highest Top-1 accuracy on three of the four datasets, including 91.61\% on the same base MagicTac-T and 91.00\% on the pneumatic marker-based PnuTac.

The performance pattern suggests that \texttt{SARL}'s representations transfer exceptionally well to other VBTS. This is because its spatially-aware objectives learn the fundamental geometry of marker displacement, a principle that generalises across different marker types. 
However, on the coating-based GelSight sensor, Barlow Twins, performed best. 
This finding suggests that while \texttt{SARL} is highly effective, the Barlow Twins feature decorrelation objective provides a slightly stronger signal to generalise to a different sensing modality. 
In conclusion, these results validate that \texttt{SARL} learns transferable, non-trivial features well-suited for a significant class of visuo-tactile sensors.

\begin{table}[t!]
\centering
\captionsetup{font=footnotesize,labelsep=period}
\vspace{-0.2cm}
\caption{Transfer learning performance on four unseen datasets. For this protocol, the pre-trained encoder was fine-tuned (only Layer 4 unfrozen). The initial SSL pre-training used the combined ViTacTip train and validation splits. All inputs are 224$\times$224. We report Top-1 and Top-5 accuracy (\%).}
\label{tab:transfer_linearprobe}
\small
\resizebox{\linewidth}{!}{
\begin{tabular}{l*{8}{c}}
\hline
& \multicolumn{2}{c}{\textbf{MagicTac-T}} &
  \multicolumn{2}{c}{\textbf{MagicTac-D}} &
  \multicolumn{2}{c}{\textbf{GelSight}} &
  \multicolumn{2}{c}{\textbf{PnuTac}} \\
\cmidrule(lr){2-3}\cmidrule(lr){4-5}\cmidrule(lr){6-7}\cmidrule(lr){8-9}
\textbf{Method} &
\textbf{Top-1} & \textbf{Top-5} &
\textbf{Top-1} & \textbf{Top-5} &
\textbf{Top-1} & \textbf{Top-5} &
\textbf{Top-1} & \textbf{Top-5} \\
\hline
SimCLR         & 80.19 & 95.38 & 77.89 & \underline{95.24} & 85.89 & 94.54 & 79.65 & 93.78 \\
BYOL           & 87.04 & 98.61 & 69.58 & 91.05 & 88.46 & 96.78 & 88.08 & \textbf{97.57} \\
Barlow Twins   & \underline{90.85} & \underline{98.86} & \underline{78.38} & 94.24 & \textbf{92.30} & \textbf{99.12} & \underline{88.98} & 97.23 \\
\texttt{SARL} (ours)  & \textbf{91.61} & \textbf{98.88} & \textbf{80.48} & \textbf{96.40} & \underline{90.24} & \underline{98.54} & \textbf{91.00} & \underline{97.32} \\
\hline
\end{tabular}}

% \vspace{2pt}
{\footnotesize Best result in each column is \textbf{bold}; second-best is \underline{underlined}.}
\vspace{-0.6cm}
\end{table}

\subsection{Ablation Studies}

% To validate our specific design choices and understand the source of performance gains, we conducted two sets of ablation studies. The first dissects the contribution of each spatial loss component, while the second quantifies the benefit of using fused multimodal data.
To verify our design choices and locate performance gains, we ran two ablations: one on spatial loss components and one on modality fusion.
Table~\ref{tab:loss_ablation} shows that the full \texttt{SARL} model, incorporating all three spatial losses, performs best, indicating strong complementarity between them. Each individual loss outperforms the global-only baseline, although their contributions vary by task. PPDA, which enforces semantic part consistency, provides the strongest single-loss results: 90.35\% for shape classification and 0.7940 MAE for edge-pose regression. RAM, which preserves geometric structure, is most effective when combined with PPDA, reducing the pose error to 0.6631.

% The results, summarised in Table~\ref{tab:loss_ablation}, confirm that the complete \texttt{SARL} model with all three spatial losses achieves the best performance, demonstrating a powerful synergy between the components. While individual losses provide improvements over a global-only baseline, their contributions appear complementary and task-dependent. The PPDA loss, which enforces semantic part consistency, provides a strong foundational improvement, individually achieving the best single-loss performance on both shape classification (90.35\%) and edge pose regression (0.7940 MAE). The RAM loss, which preserves geometric structure, shows its value most clearly when combined with PPDA, further reducing the average pose error to 0.6631.

The SAL loss acts as a crucial binding agent. While its individual contribution is more modest, its inclusion in the full model unlocks the largest performance gains. This is particularly evident in the dramatic improvement in edge pose regression, where adding SAL to the best two-loss pair reduces the MAE from 0.6570 down to 0.3955. This suggests that SAL effectively forces the model to focus on the most salient object regions, allowing the PPDA and RAM losses to operate on the most informative features. The key takeaway is that the losses are not redundant but synergistic: SAL identifies where to look, PPDA identifies what parts are present, and RAM preserves their geometric arrangement, with all three being necessary for the most robust and spatially-aware representation.

\begin{table}[t!]
\centering
\captionsetup{font=footnotesize,labelsep=period}
\vspace{-0.2cm}
\caption{Ablation of \texttt{SARL}'s spatial loss components, evaluated via a frozen-encoder linear probe on the Shape and Edge Pose tasks. Pre-training for each variant used the unlabeled ViTacTip train and validation splits. All inputs are 224$\times$224. Metrics are Top-1/Top-5 accuracy (\%) for Shape and average MAE for Edge Pose (mm, degrees).}
\label{tab:loss_ablation}
\footnotesize
\begin{tabular}{lcc}
\hline
& \multicolumn{1}{c}{\textbf{Shape}} & \multicolumn{1}{c}{\textbf{Edge Pose}} \\
\cmidrule(lr){2-2}\cmidrule(lr){3-3}
\textbf{Loss} & \textbf{Top-1 (\(\uparrow\))} & \textbf{Avg.\ MAE (\(\downarrow\))} \\
\hline
+ SAL                 & 89.41 & 0.8605 \\
+ PPDA                & 90.35 & 0.7940 \\
+ RAM                 & 90.32 & 0.8647 \\
+ SAL + PPDA          & 91.36 & 0.6570 \\
+ SAL + RAM           & 92.07 & 0.7220 \\
+ PPDA + RAM          & 91.25 & 0.6631 \\
+ SAL + PPDA + RAM    & \textbf{92.51} & \textbf{0.3955} \\
\hline
\end{tabular}

% \vspace{2pt}
{\footnotesize Best result in each column is \textbf{bold}.}
\vspace{-0.2cm}
\end{table}

\begin{table}[t!]
\centering
\captionsetup{font=footnotesize,labelsep=period}
\caption{Ablation of input data modality, evaluated via a frozen-encoder linear probe. Models were pre-trained on the unlabeled train and validation splits of each respective dataset (fused ViTacTip, visual-only ViTac, marker-only TacTip). All inputs are 224x224. We report Top-1 and Top-5 accuracy (\%) for the three classification tasks.}
\label{tab:dataablation}
\footnotesize
\begin{tabular}{lcccccc}
\hline
 & \multicolumn{2}{c}{\textbf{Shape}} & \multicolumn{2}{c}{\textbf{Grating}} & \multicolumn{2}{c}{\textbf{Material}} \\
\cmidrule(lr){2-3}\cmidrule(lr){4-5}\cmidrule(lr){6-7}
\textbf{Dataset} & \textbf{Top-1} & \textbf{Top-5} & \textbf{Top-1} & \textbf{Top-5} & \textbf{Top-1} & \textbf{Top-5} \\
\hline
TacTip    & 44.25 & 54.41 & 46.58 & 59.83 & 52.00 & 63.38 \\
ViTac     & 89.78 & \textbf{100.00} & 98.44 & \textbf{100.00} & 96.76 & 98.89 \\
ViTacTip  & \textbf{92.51} & \textbf{100.00} & \textbf{99.61} & \textbf{100.00}    & \textbf{99.27} & \textbf{99.77} \\
\hline
\end{tabular}

% \vspace{2pt}
{\footnotesize Best result in each column is \textbf{bold}.}
\vspace{-0.6cm}
\end{table}

% The results of modality ablation study presented in Table~\ref{tab:dataablation} reveals a clear performance trend. While the visual-only (ViTac) and fused multimodal (ViTacTip) data perform comparably on Top-5 accuracy, the fused ViTacTip data consistently achieves the highest Top-1 accuracy across all tasks. Both vision-based modalities are, in turn, vastly superior to the tactile-only (TacTip) data. While the visual modality is clearly dominant for these tasks, providing rich, high-resolution context, the results show that it is not sufficient on its own. The tactile data, despite performing poorly in isolation, provides a critical complementary signal when fused. It appears to act as a powerful regularizer, grounding the visual features in physical interaction and helping to resolve visual ambiguities related to fine textures and contact states. This synergy provides strong empirical evidence that hardware-fused data contains a richer and more informative learning signal than either modality alone.

The modality ablation in Table~\ref{tab:dataablation} shows a clear trend. Visual-only (ViTac) and fused (ViTacTip) are similar on Top-5, but ViTacTip consistently achieves the best Top-1 across tasks, and both vision-based inputs far outperform tactile-only (TacTip). 
While vision dominates, it is not sufficient: tactile signals, weak in isolation, regularise and ground visual features, resolving ambiguities in fine textures and contact states. This synergy supports that fused data provides a richer learning signal than either modality alone.

% \begin{table}[h]
% \caption{An Example of a Table}
% \label{table_example}
% \begin{center}
% \begin{tabular}{|c||c|}
% \hline
% One & Two\\
% \hline
% Three & Four\\
% \hline
% \end{tabular}
% \end{center}
% \end{table}

%    \begin{figure}[thpb]
%       \centering
%       \framebox{\parbox{3in}{We suggest that you use a text box to insert a graphic (which is ideally a 300 dpi TIFF or EPS file, with all fonts embedded) because, in an document, this method is somewhat more stable than directly inserting a picture.
% }}
%       %\includegraphics[scale=1.0]{figurefile}
%       \caption{Inductance of oscillation winding on amorphous
%        magnetic core versus DC bias magnetic field}
%       \label{figurelabel}
%    \end{figure}

\subsection{Discussion}
The results support our hypothesis that enforcing spatial consistency yields superior representations for robotic manipulation. 
\texttt{SARL} leads on geometrically sensitive tasks (e.g., edge pose regression), indicating that standard SSL’s collapse of spatial feature maps creates an information bottleneck. 
By acting directly on intermediate maps, the auxiliary losses (SAL, PPDA, RAM) preserve the fine-grained spatial and structural cues essential to these tasks. Ablations show the losses are complementary, not redundant, with each contributing to a stronger final representation.
% The empirical results provide strong support for our hypothesis that explicitly enforcing spatial consistency yields superior representations for robotic manipulation. \texttt{SARL}'s leading performance, especially on geometrically sensitive tasks like edge pose regression, validates our approach. Standard SSL methods create an information bottleneck by collapsing spatial feature maps. \texttt{SARL}'s success stems from its auxiliary losses (SAL, PPDA, RAM) which operate directly on these maps, compelling the model to preserve the fine-grained spatial and structural information critical for these tasks. The component ablation study demonstrates the synergistic, non-redundant nature of our proposed losses, where each contributes to a more comprehensive and robust final representation.

Despite these strengths, limitations remain. Fine-tuning spatially aware models can be fragile: the same features that make \texttt{SARL} effective are prone to catastrophic forgetting, requiring careful regularisation and learning-rate scheduling. Reliance on mid-level features also constrains transfer; fully exploiting them on out-of-domain data may require unfreezing and adapting those extractors.
% Despite its strengths, our study has limitations. The fine-tuning process for spatially-aware models can be fragile, as the very features that make \texttt{SARL} effective are susceptible to catastrophic forgetting. This necessitates careful regularisation and learning rate scheduling. Furthermore, the reliance on mid-level features for our spatial objectives introduces a specific constraint during transfer learning; unlocking their full potential on out-of-domain data may require unfreezing and adapting these critical mid-level feature extractors.

%Future work will explore integrating \texttt{SARL}'s spatial objectives into other SSL families, such as contrastive frameworks, to investigate the interplay between different collapse-prevention mechanisms. We also plan to scale the pre-training data to create a more universal visuo-tactile foundational model. A more fundamental question is whether a sufficiently strong set of spatial constraints could prevent representational collapse on its own, potentially removing the need for a global objective entirely.

\section{Conclusions}
We presented \texttt{SARL}, a spatially aware self-supervised framework for fused visuo–tactile images. 
\texttt{SARL} augments a non-contrastive backbone with three map-level objectives, including SAL, PPDA, and RAM, that preserve attentional, semantic-part, and geometric structure. 
Across six downstream tasks and nine strong SSL baselines, \texttt{SARL} delivers consistent gains, indicating that contact-rich manipulation benefits from learning signals that promote structured spatial equivariance rather than pure global invariance. Our results further suggest that hardware-fused, pixel-aligned vision–touch provides a stronger supervisory signal than either modality alone; objectives that respect this physical alignment yield transferable, manipulation-ready features.

\bibliographystyle{IEEEtran}
\bibliography{main}

@inproceedings{simclr,
  title={A simple framework for contrastive learning of visual representations},
  author={Chen, Ting and Kornblith, Simon and Norouzi, Mohammad and Hinton, Geoffrey},
  booktitle={International conference on machine learning},
  pages={1597--1607},
  year={2020},
  organization={PMLR}
}

@inproceedings{lin2023attention,
  title={Attention for robot touch: Tactile saliency prediction for robust sim-to-real tactile control},
  author={Lin, Yijiong and Comi, Mauro and Church, Alex and Zhang, Dandan and Lepora, Nathan F},
  booktitle={2023 IEEE/RSJ International Conference on Intelligent Robots and Systems (IROS)},
  pages={10806--10812},
  year={2023},
  organization={IEEE}
}

@inproceedings{moco,
  title={Momentum contrast for unsupervised visual representation learning},
  author={He, Kaiming and Fan, Haoqi and Wu, Yuxin and Xie, Saining and Girshick, Ross},
  booktitle={Proceedings of the IEEE/CVF conference on computer vision and pattern recognition},
  pages={9729--9738},
  year={2020}
}

@article{byol,
  title={Bootstrap your own latent-a new approach to self-supervised learning},
  author={Grill, Jean-Bastien and Strub, Florian and Altch{\'e}, Florent and Tallec, Corentin and Richemond, Pierre and Buchatskaya, Elena and Doersch, Carl and Avila Pires, Bernardo and Guo, Zhaohan and Gheshlaghi Azar, Mohammad and others},
  journal={Advances in neural information processing systems},
  volume={33},
  pages={21271--21284},
  year={2020}
}

@inproceedings{simsiam,
  title={Exploring simple siamese representation learning},
  author={Chen, Xinlei and He, Kaiming},
  booktitle={Proceedings of the IEEE/CVF conference on computer vision and pattern recognition},
  pages={15750--15758},
  year={2021}
}

@inproceedings{barlow,
  title={Barlow twins: Self-supervised learning via redundancy reduction},
  author={Zbontar, Jure and Jing, Li and Misra, Ishan and LeCun, Yann and Deny, St{\'e}phane},
  booktitle={International conference on machine learning},
  pages={12310--12320},
  year={2021},
  organization={PMLR}
}

@inproceedings{vicreg,
  title={VICReg: Variance-Invariance-Covariance Regularization For Self-Supervised Learning},
  author={Bardes, Adrien and Ponce, Jean and Lecun, Yann},
  booktitle={ICLR 2022-International Conference on Learning Representations},
  year={2022}
}

@article{swav,
  title={Unsupervised learning of visual features by contrasting cluster assignments},
  author={Caron, Mathilde and Misra, Ishan and Mairal, Julien and Goyal, Priya and Bojanowski, Piotr and Joulin, Armand},
  journal={Advances in neural information processing systems},
  volume={33},
  pages={9912--9924},
  year={2020}
}

@article{densecl,
  title={Densecl: A simple framework for self-supervised dense visual pre-training},
  author={Wang, Xinlong and Zhang, Rufeng and Shen, Chunhua and Kong, Tao},
  journal={Visual Informatics},
  volume={7},
  number={1},
  pages={30--40},
  year={2023},
  publisher={Elsevier}
}

@inproceedings{noroozi2016unsupervised,
  title={Unsupervised learning of visual representations by solving jigsaw puzzles},
  author={Noroozi, Mehdi and Favaro, Paolo},
  booktitle={European conference on computer vision},
  pages={69--84},
  year={2016},
  organization={Springer}
}

@inproceedings{imgrot,
  title={Unsupervised representation learning by predicting image rotations},
  author={Komodakis, Nikos and Gidaris, Spyros},
  booktitle={International Conference on Learning Representations (ICLR)},
  year={2018}
}

@article{zhang2024tacpalm,
  title={Tacpalm: A soft gripper with a biomimetic optical tactile palm for stable precise grasping},
  author={Zhang, Xuyang and Yang, Tianqi and Zhang, Dandan and Lepora, Nathan F},
  journal={IEEE Sensors Journal},
  year={2024},
  publisher={IEEE}
}

@article{babadian2023fusion,
  title={Fusion of tactile and visual information in deep learning models for object recognition},
  author={Babadian, Reza Pebdani and Faez, Karim and Amiri, Mahmood and Falotico, Egidio},
  journal={Information Fusion},
  volume={92},
  pages={313--325},
  year={2023},
  publisher={Elsevier}
}

@inproceedings{touch,
  title={Touch and go: learning from human-collected vision and touch},
  author={Yang, Fengyu and Ma, Chenyang and Zhang, Jiacheng and Zhu, Jing and Yuan, Wenzhen and Owens, Andrew},
  booktitle={Proceedings of the 36th International Conference on Neural Information Processing Systems},
  pages={8081--8103},
  year={2022}
}

@inproceedings{anytouch,
  title={AnyTouch: Learning Unified Static-Dynamic Representation across Multiple Visuo-tactile Sensors},
  author={Feng, Ruoxuan and Hu, Jiangyu and Xia, Wenke and Shen, Ao and Sun, Yuhao and Fang, Bin and Hu, Di and others},
  booktitle={The Thirteenth International Conference on Learning Representations},
  year={2025}
}

@inproceedings{zhang2024self,
  title={Self-supervised bayesian visual imitation learning applied to robotic pouring},
  author={Zhang, Dan-Dan and Zheng, Yu and Fan, Wen and Lepora, Nathan and Zhang, Zhengyou},
  booktitle={2024 IEEE International Conference on Industrial Technology (ICIT)},
  pages={1--7},
  year={2024},
  organization={IEEE}
}

@inproceedings{mvitactip,
  title={Multimodal visual-tactile representation learning through self-supervised contrastive pre-training},
  author={Dave, Vedant and Lygerakis, Fotios and Rueckert, Elmar},
  booktitle={2024 IEEE International Conference on Robotics and Automation (ICRA)},
  pages={8013--8020},
  year={2024},
  organization={IEEE}
}

@article{fan2402vitactip,
  title={ViTacTip: Design and verification of a novel biomimetic physical vision-tactile fusion sensor. arXiv 2024},
  author={Fan, W and Li, H and Si, W and Luo, S and Lepora, N and Zhang, D},
  journal={arXiv preprint arXiv:2402.00199}
}

@inproceedings{t3,
  title={Transferable Tactile Transformers for Representation Learning Across Diverse Sensors and Tasks},
  author={Zhao, Jialiang and Ma, Yuxiang and Wang, Lirui and Adelson, Edward},
  booktitle={Conference on Robot Learning},
  pages={3766--3779},
  year={2025},
  organization={PMLR}
}

@article{uniT,
  title={UniT: Data Efficient Tactile Representation With Generalization to Unseen Objects},
  author={Xu, Zhengtong and Uppuluri, Raghava and Zhang, Xinwei and Fitch, Cael and Crandall, Philip Glen and Shou, Wan and Wang, Dongyi and She, Yu},
  journal={IEEE Robotics and Automation Letters},
  volume={10},
  number={6},
  year={2025},
  publisher={IEEE}
}

@inproceedings{upvital,
  title={UpViTaL: Unpaired Visual-Tactile Self-Supervised Representation Learning for Dexterous Robotic Manipulation},
  author={Han, Guwen and Liu, Qingtao and Cui, Yu and Chen, Anjun and Chen, Jiming and Ye, Qi},
  booktitle={2025 IEEE International Conference on Robotics and Automation (ICRA)},
  pages={11838--11844},
  year={2025},
  organization={IEEE}
}

@article{GelStereo,
  title={Self-supervised contact geometry learning by GelStereo visuotactile sensing},
  author={Cui, Shaowei and Wang, Rui and Hu, Jingyi and Zhang, Chaofan and Chen, Lipeng and Wang, Shuo},
  journal={IEEE Transactions on Instrumentation and Measurement},
  volume={71},
  pages={1--9},
  year={2021},
  publisher={IEEE}
}

@inproceedings{sparsh,
  title={Sparsh: Self-supervised touch representations for vision-based tactile sensing},
  author={Higuera, Carolina and Sharma, Akash and Bodduluri, Chaithanya Krishna and Fan, Taosha and Lancaster, Patrick and Kalakrishnan, Mrinal and Kaess, Michael and Boots, Byron and Lambeta, Mike and Wu, Tingfan and others},
  booktitle={Conference on Robot Learning},
  pages={885--915},
  year={2025},
  organization={PMLR}
}

@article{vitactip,
  title={Design and Benchmarking of a Multimodality Sensor for Robotic Manipulation With GAN-Based Cross-Modality Interpretation},
  author={Zhang, Dandan and Fan, Wen and Lin, Jialin and Li, Haoran and Cong, Qingzheng and Liu, Weiru and Lepora, Nathan F and Luo, Shan},
  journal={IEEE Transactions on Robotics},
  volume={41},
  pages={1278--1295},
  year={2025},
  publisher={IEEE}
}

@inproceedings{pnutac,
  title={PnuTac: A vision-based pneumatic tactile sensor for slip detection and object classification},
  author={Rayamane, Prasad and Herbert, Peter and Munguia-Galeano, Francisco and Ji, Ze},
  booktitle={2023 29th International Conference on Mechatronics and Machine Vision in Practice (M2VIP)},
  pages={1--6},
  year={2023},
  organization={IEEE}
}

@inproceedings{magictac,
  title={Magictac: A novel high-resolution 3d multi-layer grid-based tactile sensor},
  author={Fan, Wen and Li, Haoran and Zhang, Dandan},
  booktitle={2024 IEEE International Conference on Robotics and Automation (ICRA)},
  pages={388--394},
  year={2024},
  organization={IEEE}
}

@article{minc,
  title={Representation Learning via Non-Contrastive Mutual Information},
  author={Guo, Zhaohan Daniel and Pires, Bernardo Avila and Khetarpal, Khimya and Schuurmans, Dale and Dai, Bo},
  journal={arXiv preprint arXiv:2504.16667},
  year={2025}
}

@inproceedings{cplearn,
  title={Collapse-Proof Non-Contrastive Self-Supervised Learning},
  author={Sansone, Emanuele and Lebailly, Tim and Tuytelaars, Tinne},
  booktitle={Forty-second International Conference on Machine Learning, 2025},
  year={2025}
}

@inproceedings{resnet,
  title={Deep residual learning for image recognition},
  author={He, Kaiming and Zhang, Xiangyu and Ren, Shaoqing and Sun, Jian},
  booktitle={Proceedings of the IEEE conference on computer vision and pattern recognition},
  pages={770--778},
  year={2016}
}

@article{ssl,
  title={Self-supervised representation learning: Introduction, advances, and challenges},
  author={Ericsson, Linus and Gouk, Henry and Loy, Chen Change and Hospedales, Timothy M},
  journal={IEEE Signal Processing Magazine},
  volume={39},
  number={3},
  pages={42--62},
  year={2022},
  publisher={IEEE}
}

@article{sslsurvey,
  title={A survey on self-supervised representation learning},
  author={Uelwer, Tobias and Robine, Jan and Wagner, Stefan Sylvius and H{\"o}ftmann, Marc and Upschulte, Eric and Konietzny, Sebastian and Behrendt, Maike and Harmeling, Stefan},
  journal={arXiv preprint arXiv:2308.11455},
  year={2023}
}

@article{jea,
  title={A path towards autonomous machine intelligence version 0.9. 2, 2022-06-27},
  author={LeCun, Yann},
  journal={Open Review},
  volume={62},
  number={1},
  pages={1--62},
  year={2022}
}

@inproceedings{gao2016deep,
  title={Deep learning for tactile understanding from visual and haptic data},
  author={Gao, Yang and Hendricks, Lisa Anne and Kuchenbecker, Katherine J and Darrell, Trevor},
  booktitle={2016 IEEE international conference on robotics and automation (ICRA)},
  pages={536--543},
  year={2016},
  organization={IEEE}
}

@article{fan2025crystaltac,
  title={CrystalTac: Vision-Based Tactile Sensor Family Fabricated via Rapid Monolithic Manufacturing},
  author={Fan, Wen and Li, Haoran and Zhang, Dandan},
  journal={Cyborg and Bionic Systems},
  volume={6},
  pages={0231},
  year={2025},
  publisher={AAAS}
}

@article{fan2023tac,
  title={Tac-vgnn: A voronoi graph neural network for pose-based tactile servoing},
  author={Fan, Wen and Yang, Max and Xing, Yifan and Lepora, Nathan F and Zhang, Dandan},
  journal={arXiv preprint arXiv:2303.02708},
  year={2023}
}

\end{document}